\documentclass[journal]{IEEEtran}
\usepackage{xcolor,soul,framed} %,caption

\colorlet{shadecolor}{yellow}
\usepackage[pdftex]{graphicx}
\graphicspath{{../pdf/}{../jpeg/}}
\DeclareGraphicsExtensions{.pdf,.jpeg,.png}

\usepackage[cmex10]{amsmath}
\usepackage{algorithm}
\usepackage{algorithmic}

\usepackage{pdfpages}
\usepackage{epsfig}
\usepackage{graphics}
\usepackage{float}
\usepackage{hyperref}
\usepackage{eqparbox}

\usepackage{csquotes}% advanced facilities for inline and display quotations
\usepackage{subfigure}
\usepackage{flushend}
\newtheorem{assum}{Assumption}
\usepackage{mathtools}

\usepackage{cite}
\usepackage{tablefootnote}
\usepackage{array}
\usepackage{mdwmath}
\usepackage{amssymb}
\usepackage{amsmath}
\usepackage{mdwtab}
\usepackage{eqparbox}
\usepackage{url}
\usepackage{xcolor}
\usepackage{comment}
\usepackage{footnote}
\usepackage{multirow}
\usepackage{graphicx}
\usepackage{caption}
\usepackage{placeins}

\usepackage{array, makecell}

\usepackage{tikz}

\usetikzlibrary{fit,positioning}
\usetikzlibrary{bayesnet}
\usetikzlibrary{arrows}

\begin{document}
\bstctlcite{IEEEexample:BSTcontrol}
    %\title{Numerical Attributes Learning for Cardiac Failure Diagnostic and Vital Distress at Birth from Clinical Narratives - A LESA-CamemBERT-bio Approach}

    \title{Hybrid Deep Learning-Based for Enhanced Occlusion Segmentation in PICU Patient Monitoring}
    
    \author{Mario Francisco Munoz, Hoang Vu Huy, Thanh-Dung Le,~\IEEEmembership{Member,~IEEE,}\\
    Philippe Jouvet, and Rita Noumeir,~\IEEEmembership{Member,~IEEE}

 \thanks{This work was supported in part by the Natural Sciences and Engineering Research Council (NSERC), in part by the Institut de Valorisation des données de l’Université de Montréal (IVADO), and in part by the Fonds de la recherche en sante du Quebec (FRQS).}

%}
\thanks{Mario Francisco Munoz is with the Electrical Engineering Department, École de Technologie Supérieure, Montréal, QC, Canada, and with the Saint-Justine Mother and Child University Hospital Center, Montréal, QC, Canada. (e-mail: mario.munoz.hsj@ssss.gouv.qc.ca).}

\thanks{Hoang Vu Huy is with with the Biomedical Information Processing Lab, \'{E}cole de Technologie Sup\'{e}rieure, University of Qu\'{e}bec, Canada. (e-mail: hoang.vu-huy.1@ens.etsmtl.ca).}

\thanks{Thanh-Dung Le is with the Biomedical Information Processing Lab, \'{E}cole de Technologie Sup\'{e}rieure, University of Qu\'{e}bec, Canada, and with the Interdisciplinary Centre for Security, Reliability, and Trust (SnT), University of Luxembourg, Luxembourg. (e-mail: thanh-dung.le@uni.lu).}

\thanks{Philippe Jouvet is with CHU Sainte-Justine Hospital, University of Montreal, Canada. (e-mail: philippe.jouvet.med@ssss.gouv.qc.ca).}

\thanks{Rita Noumeir is with the Biomedical Information Processing Lab, \'{E}cole de Technologie Sup\'{e}rieure, University of Qu\'{e}bec, Canada. (e-mail: rita.noumeir@etsmtl.ca).}
 
}

% The paper headers
\markboth{IEEE, VOL., NO., 2024.
}{Mario Francisco Munoz \MakeLowercase{\textit{et al.}}: Hybrid Deep Learning-Based for Enhanced Occlusion Segmentation in PICU Patient Monitoring}

% ====================================================================
\maketitle

% === ABSTRACT ====================================================================
% =================================================================================
\begin{abstract}
% Remote patient monitoring (RPM) has emerged as a promising non-invasive approach, leveraging digital technologies and computer vision (CV) to replace invasive physiological monitoring. Neonatal and pediatric healthcare departments have readily embraced these remote monitoring tools due to their non-invasive nature. However, the Pediatric Intensive Care Units (PICUs) face a unique challenge – various occlusions can obstruct accurate image analysis and interpretation. Consequently, it becomes imperative to identify the presence and locations of these occlusions or, in simpler terms, to separate them from the subjects being analyzed. 
% Traditionally, deep learning-based algorithms are employed for segmentation tasks. However, they typically demand extensive training data for satisfactory performance. 
% In clinical settings, meeting this data requirement can be problematic due to patient data sensitivity, resulting in suboptimal segmentation quality. 

Remote patient monitoring has emerged as a prominent non-invasive method, using digital technologies and computer vision (CV) to replace traditional invasive monitoring. While neonatal and pediatric departments embrace this approach, Pediatric Intensive Care Units (PICUs) face the challenge of occlusions hindering accurate image analysis and interpretation. \textit{Objective}:
In this study, we propose a hybrid approach to effectively segment common occlusions encountered in remote monitoring applications within PICUs. Our approach centers on creating a deep-learning pipeline for limited training data scenarios. \textit{Methods}:
First, a combination of the well-established Google DeepLabV3+ segmentation model with the transformer-based Segment Anything Model (SAM) is devised for occlusion segmentation mask proposal and refinement. We then train and validate this pipeline using a small dataset acquired from real-world PICU settings with a Microsoft Kinect camera, achieving an Intersection-over-Union (IoU) metric of 85\%. \textit{Results}:
Both quantitative and qualitative analyses underscore the effectiveness of our proposed method. The proposed framework yields an overall classification performance with 92.5\% accuracy, 93.8\% recall, 90.3\% precision, and 92.0\% F1-score. Consequently, the proposed method consistently improves the predictions across all metrics, with an average of 2.75\% gain in performance compared to the baseline CNN-based framework. \textit{Conclusions}: Our proposed hybrid approach significantly enhances the segmentation of occlusions in remote patient monitoring within PICU settings. This advancement contributes to improving the quality of care for pediatric patients, addressing a critical need in clinical practice by ensuring more accurate and reliable remote monitoring. 

\end{abstract}

\begin{IEEEkeywords}
Computer Vision,
Data Augmentation,
Deep Learning,
Model Fusion,
Occlusions,
Pediatrics Intensive Care,
Remote Patient Monitoring (RPM),
Segmentation.
\end{IEEEkeywords}

\IEEEpeerreviewmaketitle

\textbf{\textit{Clinical and Translational Impact Statement---} }
This study presents a hybrid deep-learning approach for segmenting occlusions in remote patient monitoring within PICUs, effectively addressing limited training data challenges. Combining the Google DeepLabV3+ model with the transformer-based SAM, the method demonstrates high accuracy and reliability in real-world PICU settings. This practical application of advanced deep-learning techniques enhances image analysis and interpretation, ultimately improving the quality of pediatric care.

\section{Introduction}
\label{sec:introduction}
% CLINICAL BACKGROUND

\begin{figure}[!t]
\centerline{\includegraphics[width=0.5\columnwidth]{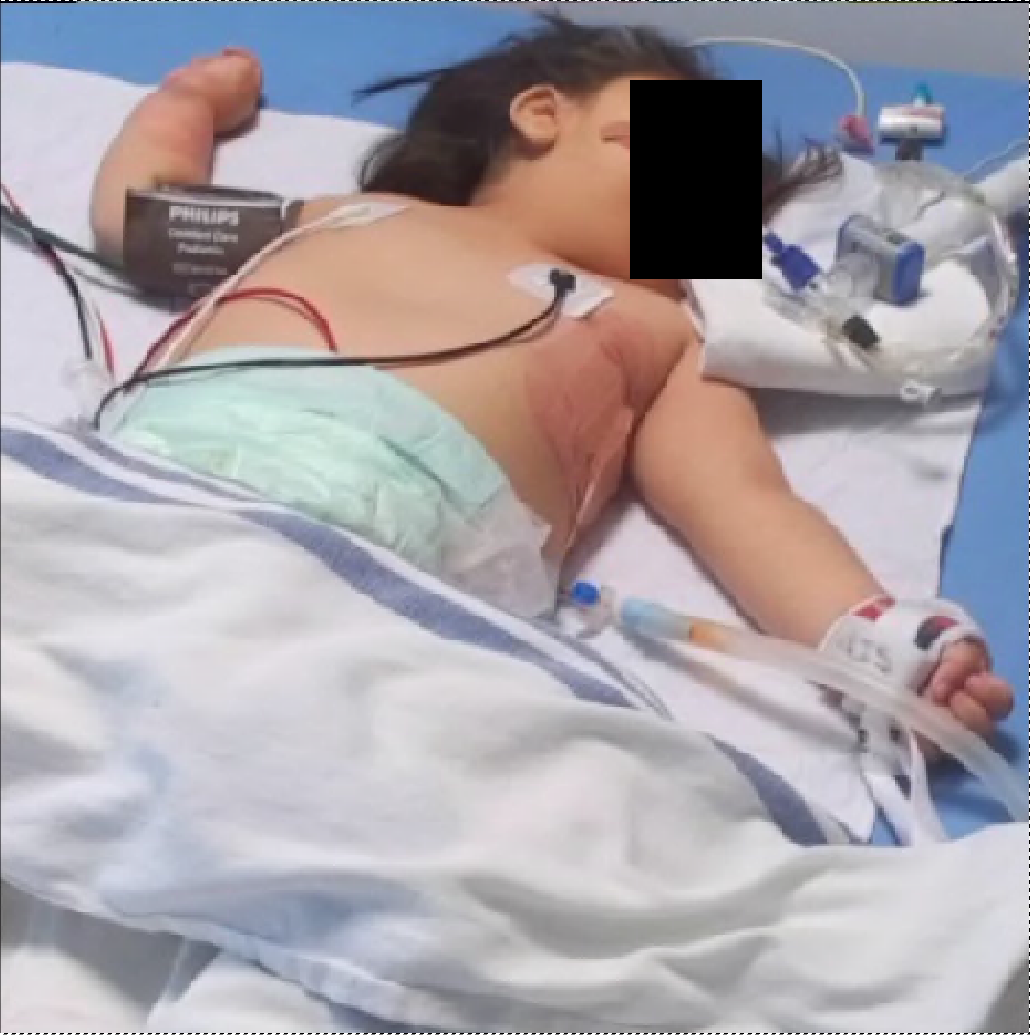}}
% \captionsetup{justification=centering}
\caption{An illustrative image of a PICU patient with different occlusions in our CHU Sainte Justine's database.}
\label{fig:occlusionexample}
\end{figure}

\IEEEPARstart{R}{emote} patient monitoring (RPM) \cite{farias2020remote} has been considered a viable alternative to invasive physiological monitoring, particularly thanks to the rapid advances in deep learning-based computer vision technologies.
Computer vision (CV) tasks like remote photoplethysmography (R-PPG)\cite{allado_remote_2022}, pose estimation \cite{andriluka_2d_2014}, and thermal circulation monitoring systems \cite{shcherbakova2021optical} provide important physiological data about patients without encumbering them.
RPM is particularly important in pediatrics due to its potential to improve access to care, enhance chronic disease management, reduce hospitalizations, cater to pediatric-specific monitoring needs, involve parents in care, and enable early intervention and emergency response\cite{foster2022remote}.

% MOTIVATIONS AND APPLICATIONS
From the CV perspective, the performance of downstream tasks depends heavily on the quality of input data.
Noise and outliers usually prevent automatic analyses from being accurate and robust.
This is particularly critical in the case of deep learning models trained on a limited amount of data.
To increase the training effectiveness, irrelevant information needs to be singled out and treated appropriately. 
For example, R-PPG methods rely on the change in the amount of light reflected on the skin driven by blood streams underneath. 
If occlusions such as tubes and breathers are mistakenly included in the region of interest, estimation accuracy might be negatively impacted \cite{kossack2022perfusion}.
Another example is the thermography-based approach \cite{shcherbakova2021optical} for estimating the patient's cardiac output. 
Obviously, this method cannot be automated without a way to segment the occlusions in the image, as they clearly alter the thermal readings of the camera. 
Ironically, CV tasks in PICU settings whose reliability and accuracy are most prioritized actually suffer the most because of the variety and large quantity of occlusions (Figure \ref{fig:occlusionexample}).

%\textbf{Please ask Dr. Jouvet for the clinical justification that highlights why we need this work in our PICU. Describing the practical clinical implications at CHU Sainte Justine (CHUSJ) from this work makes our work more convincible and applicable to the PICU at CHUSJ.}

For example, algorithms are developed to automatically measure the temperature gradient between the forehead and toe to assess hemodynamic status in children \cite{bridier2023hemodynamic}.
If there is a plaster or any other occlusion on the forehead, temperature measurement will not be accurate. 
As temperature distribution is due to blood flow distribution, other body parts are under investigation to determine if they are relevant for hemodynamic assessment \cite{shcherbakova2021optical}.
Another example is the pain assessment on video analyzing facial mimic recognition patterns. 
In the case of nasogastric tubes or other occlusion, the deep learning models developed are not accurate, and automatic real-time analysis can not be performed \cite{prinsen2021automatic}.

These aforementioned difficulties can be overcome when irrelevant information due to occlusions is properly treated.
In CV, this can be attained by the localization and isolation of occlusions, or equivalently occlusion segmentation \cite{yuan2021robust}.
When the distinction of individual occlusions is not necessary, semantic segmentation is a usual choice.
Conventionally, semantic segmentation is the action of grouping an input image's pixels into different classes, each of which corresponds to a semantic label, e.g. "car" or "person"\cite{guo2018review}.
Among existing methods for semantic segmentation, deep learning-based approaches are the most popular with a wide range of applications in reality.

The high accuracy and robustness of the most successful deep learning models are usually attributed to the availability of large training datasets which might comprise millions of images.
However, this is typically not the case in PICU settings where images of patients can only be taken and used on strict conditions including parents' consent.  
Image quality is also a problem due to non-ideal light conditions and noises. 
In this scenario, transfer learning can be employed as a form of prior knowledge that is acquired from huge datasets.
Depending on the nature of the source and the target deep-learning tasks, the transferred parts of the source model can continue to be trained on the target domain (fine-tuning) or can be used as is (zero-shot learning). 
While transfer learning has proven its pivotal role in the contemporary deep learning community, gaps between target medical domains and source domains on which deep learning models are pre-trained can affect their zero-shot learning performance (\cite{alzubaidi2020towards, niu2021distant}). 
Thus, there is an urgent need to develop segmentation algorithms that can perform robustly in this special environment with limited annotation and transferability.

\begin{figure*}[!h]
\centerline{\includegraphics[width=0.9\textwidth]{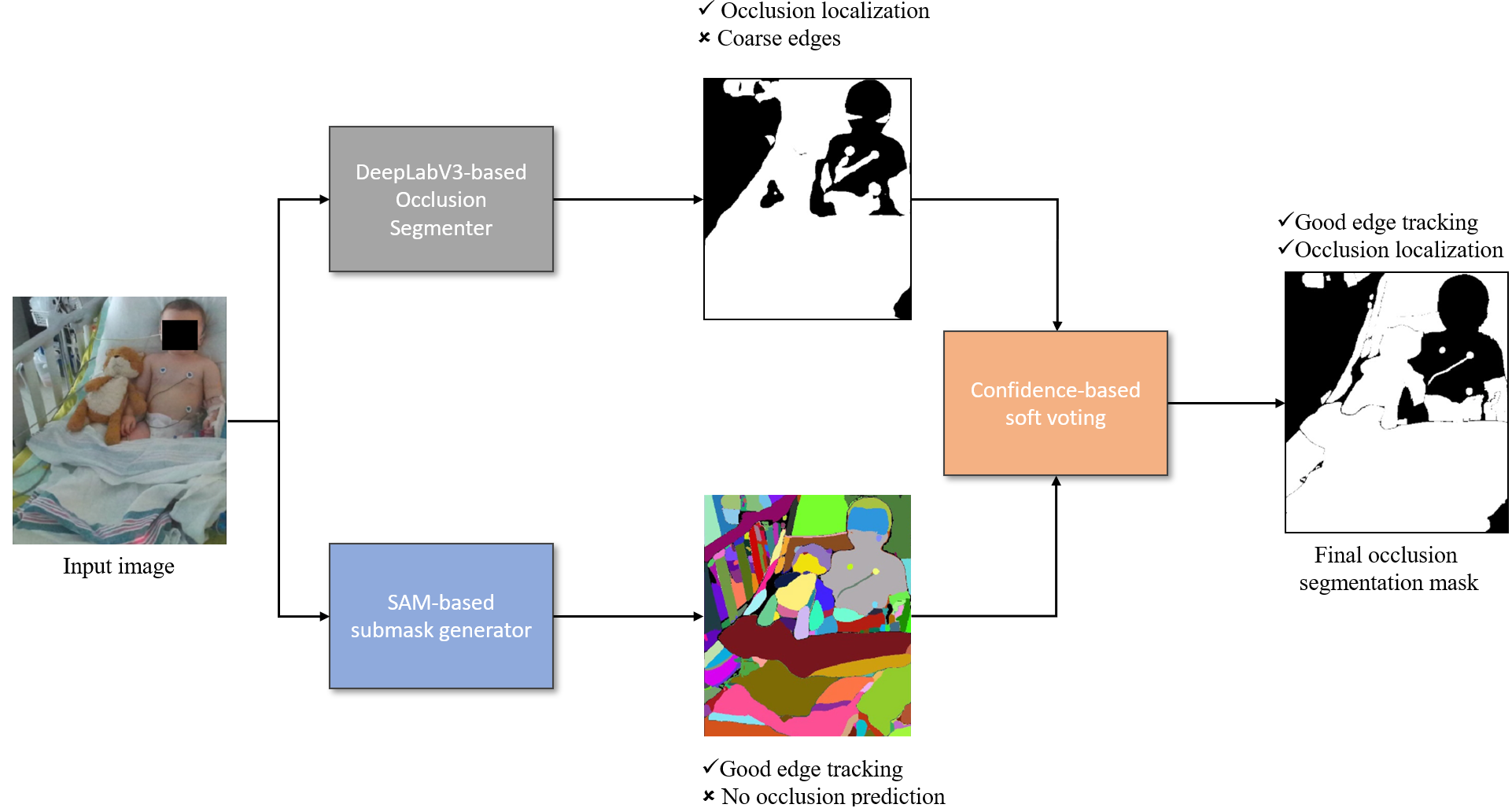}}
% \captionsetup{justification=centering}
\caption{Our proposed pipeline (SOSS) for occlusion segmentation. 
The input image is fed to both the top branch and bottom branch simultaneously.
Top branch: our DeepLabV3+-based network segments the input image and produces a semantic (occlusion) mask proposal. Bottom branch: the SAM-based generator produces a partitioning of the input image without corresponding semantic labels. Both kinds of masks are then fused using our proposed confidence-based soft voting mechanism for the final occlusion segmentation mask. This aims to add semantic information to the SAM branch while simultaneously improving the segmentation quality of the DeepLab branch.}
\label{fig:pipeline}
\end{figure*}

Understanding these challenges as well as being motivated by the necessity of occlusion processing in RPM applications, in this work we proposed a novel deep learning-based framework for occlusion segmentation in PICU settings.
Our contributions include:
\begin{itemize}
    \item First, a real-world dataset of pediatric patients' images at CHU Sainte Justine Hospital (CHSJ) with occlusions of various kinds, shapes, and sizes has been collected and annotated, which can readily be used for training and evaluating different algorithms on occlusion detection and segmentation tasks.
    \item Second, a convolutional neural network (CNN)-based model is trained and evaluated on the dataset, reporting plausible performance and generalization for occlusion segmentation in PICU settings where limited and unbalanced training data is a challenging problem.
    \item Third, a novel data-efficient fusion pipeline named SOSS (SAM-powered Occlusion Segmentation via Soft-voting) is introduced (Figure \ref{fig:pipeline}).
    This pipeline leverages a foundation-class transformer-based image segmentation model as a means to refine the output of the preceding CNN-based occlusion segmentation model, effectively proving its capability of segmenting occlusions of various kinds in our considered clinical use case.
\end{itemize}
With this novel framework, we intend to promote the applicability of RPM for PICU deployments and other real-world uses of computer vision.

\section{Related works}
In this section, we summarize related works in learning-based semantic segmentation, segmentation with Segment-Anything-Model (SAM), and deep-learning for occlusion handling.

\subsection{Convolutional Neural Networks for Semantic Segmentation}
Convolutional Neural Networks (CNN)-based models have been extensively used for semantic segmentation tasks in particular, as well as for other image processing tasks in general.
This branch of deep-learning algorithms centers around the convolution operation which acts on different parts of an image, resulting in hierarchical structures that gradually extract semantic and spatial information at different scales.
Fully Convolutional Network (FCN) \cite{long2015fully} was among the first attempts to harness convolutions more efficiently for dense prediction tasks such as segmentation.
Around the same time, a novel segmentation model named U-Net\cite{ronneberger2015u} introduced a symmetric encoder-decoder architecture with the notion of skip layers, preserving detailed information during the up-sampling branches and proving to be adopted widely in biomedical applications \cite{zhou2019unet++,huang2020unet,krithika2022review}.
Another well-known segmentation model is DeepLab \cite{chen2017deeplab} which proposed an enhanced processing operation named atrous (dilated) convolution as a solution to the problem of limited receptive field in previous CNN-based algorithms without using inflating the model parameter number too much.
This novel operation was particularly effective for semantic information mining and thus was suitable for semantic segmentation tasks.

The latest version of DeepLab, the DeepLabV3+ model\cite{chen2017rethinking} was the combination of the atrous convolution, a multi-scale feature fusion named ASPP, and the overall encoder-decoder architecture from U-Net.
DeepLabV3+ outperformed other models on both the PASCAL VOC 2012 test set and the Cityscapes dataset.
Overall, CNN-based segmentation algorithms are still widely used nowadays with the clear advantage of being data-efficient and more suitable for small-to-medium-scaled training datasets.
% Using the earlier CRFs at the end of a network built with residual blocks, the first DeepLab \cite{chen2017deeplab} model achieves state-of-the-art performance on the COCO segmentation \cite{lin2014microsoft} dataset, amongst many others.
% This first DeepLab model simultaneously tackles two issues.
% The first is the issue of reduced feature resolution.
% They mitigate this issue by removing some downsampling layers in exchange for atrous convolution layers which also use less parameters and increase the field of view of filters.
% The second issue is reduced localization accuracy due using shift-invariant CNNs.
% The authors mitigate this issue by 00000000000000implementing the aforementioned CRFs as refining modules at the end of the network to enhance the end predictions.
% In succession to this work but in the same paper, an improved DeepLabV2 effectively improves DeepLab performance by sensitizing the model to smaller objects through atrous spatial pyramid pooling \cite{chen2017deeplab}.
% In the same line of models, DeepLabV3 \cite{chen2017rethinking} improves upon its last version by implementing an encoder-decoder architecture that recovers spatial information in order to have sharper object boundaries.
% Lastly, the DeepLabV3+ model\cite{chen2017rethinking} adds another decoder module to refine the output segmentation masks further.

\subsection{Vision Transformers for Semantic Segmentation}
Vision transformers (ViT), a class of models that leverage self-attention mechanisms and transformer architectures, have emerged as powerful tools for computer vision tasks. 
One of the first works to apply transformers to semantic segmentation is Segmenter \cite{strudel2021segmenter}, which introduces a simple and efficient design based on the ViT architecture.
Segmenter uses the output embeddings of ViT as input to a linear or mask transformer decoder which generates segmentation masks.
Another high-recognizable work that leverages ViT for semantic segmentation is SegFormer \cite{xie2021segformer} which proposes a hierarchical transformer encoder (Mix Transformer) and a lightweight decoder.
Similar to DeepLabV3+ and U-net, SegFormer aims to capture multi-scale features with convolutional layer enhancement to reduce computational cost.
A recent work that improves transformers for semantic segmentation is Transformer Scale Gate (TSG) \cite{shi2023transformer} which introduces a novel scale gate module that dynamically adjusts the scale of features in each attention head. 
TSG aims to address the issue of scale variation in semantic segmentation, which can cause some features to be dominant or suppressed by others.
TSG can be integrated into any transformer-based model for semantic segmentation, and it improves the performance of Segmenter and SegFormer on Cityscapes and ADE20K datasets.

In summary, unlike CNNs, vision transformers are able to directly capture global context and long-range dependencies between pixels without the need for complex hierarchical systems and thus have great potential for semantic segmentation.
However, these advantages come at the cost of increased complexity and the need for large-sized training datasets.
To address the problem of data-hungry ViTs, fine-tuning pre-trained models is adopted as a common strategy.
Nevertheless, even fine-tuning cannot guarantee the generalization of pre-trained models on new domains and tasks, especially with limited training data \cite{li2022domain,hua2022fine}.

\begin{figure*}[htbp]
\centerline{\includegraphics[width=0.6\textwidth]{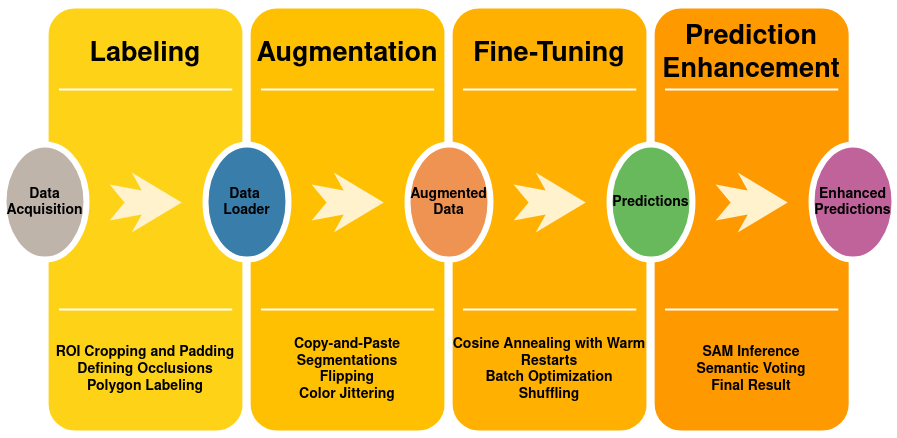}}
\caption{A summary of our workflow, including 4 steps: labeling, data augmentation, model fine-tuning, and prediction.}
\label{fig:workflow}
\end{figure*}

\subsection{Semantic Segmentation using SAM}
A notable state-of-the-art transformer model is the Segment Anything Model, or SAM \cite{kirillov2023segment}.
SAM has been trained on a large dataset of 11 million images and 1.1 billion masks, and it can transfer zero-shot to new image distributions and tasks. 
SAM is designed to be promptable, meaning that it can adapt to different user preferences and scenarios by changing the input prompts. 
SAM is also efficient, as it uses a simple and lightweight architecture based on ViTs.
One weakness of SAM is that this model is not originally dedicated to semantic segmentation tasks. 

\cite{zhang2023sam} adapted SAM for semantic segmentation by introducing trainable class prompts and a pathology encoder. 
It improved the performance of SAM on two public pathology datasets, the BCSS and the CRAG datasets.
\cite{momma2020p} used SAM as a pre-processing step to generate initial segmentation masks for LiDAR point cloud frames. It then refines the masks with a post-processing network that enforces consistency constraints between consecutive frames.
Recently, \cite{li2023semantic} proposed Semantic-SAM, a universal image segmentation model that can segment and recognize anything at any desired granularity.
They consolidate multiple datasets across three granularities and introduce decoupled classification for objects and parts. 
Overall, the advent of SAM offers a novel way to achieve state-of-the-art segmentation performance at relatively low training costs compared with previous ViT-based solutions.

\subsection{Deep-learning for Occlusion Handling and Segmentation}
There have been a number of works related to the topic of occlusion handling and segmentation in the existing literature.
Recently, deep learning has become a widely adopted approach to occlusion segmentation, particularly CNN-based ones.
However, CNNs often struggle to handle complex occlusion scenarios, such as highly overlapping objects, large occlusion ratios, and diverse occlusion patterns. 
Therefore, various techniques have been proposed to enhance the robustness and accuracy of CNN-based occlusion segmentation methods.
For example, \cite{ke2021deep} proposed a bilayer architecture named BCNet to model the occlusion relationship between the occluding and occluded objects.
The method can decouple the boundaries of both the occluding and occluded instances and consider their interaction during mask regression.
In \cite{yuan2021robust}, the authors utilized an occluder segmentation network as part of their intricate occlusion reasoning module (ORM) for probabilistic modeling segmentation error, ultimately recovering the object orders which is beneficial for multi-object occlusion cases.
The method can supposedly enhance the robustness of deep learning algorithms for instance segmentation tasks.
Similarly, \cite{lazarow2020learning} developed a deep neural network to model how instance masks should overlap each other as a binary relation, arguing that common fusion of instance segmentation map with confidence score alone cannot sufficiently model the natural occlusion ordering.
Finally, the more recent work \cite{back2022unseen} opted to handle occluded objects and the problem of unseen object amodal instance segmentation via hierarchical occlusion modeling, which focuses also on feature fusion to recover the occluded region (amodal segmentation) and predict the occlusion order.

\section{Methodology}
\label{sec:methodology}

In this work, we define occlusions as common objects in the PICU patient's proximity that may impede a direct view of a patient's body. 
Walls, floors, and ceilings as well as other background objects not usually found in the same bed as the patient are labeled as non-occlusions.
Due to the wide variety in the pose of the patient, in many cases, it is difficult to tell where the obscured body parts lie underneath the occlusions.
Therefore, occlusion objects such as blankets or tubes lying beside the patient might also be labeled as occlusions.

A simplified workflow for this part is presented in Figure \ref{fig:workflow}.

Namely, our work comprises 4 main stages besides data acquisition: 
\begin{itemize}
    \item Labeling: preprocessing (ROI cropping and padding), occlusion defining, and segmenting in polygon format.
    \item Data augmentation: applying mild augmentations (flipping, color jittering, random brightness \& contrast) and strong augmentations(Copy-Paste).
    \item Model fine-tuning: Semantic segmentation modeling, fine-tuning with cosine annealing with warm restarts, batch optimization, and shuffling.
    \item Prediction enhancement: SAM-based image segmentation, semantic soft-voting, and producing final occlusion segmentation masks.
\end{itemize}

\subsection{Data Acquisition}
The study was approved by the research ethic board (REB) of Sainte-Justine hospital (project number 2020-2287) and was conducted on a video database approved by the same REB (database project number  2016-1242).

175 different high-resolution color digital photographs from consenting patients were acquired in the intensive care unit at Saint Justine's Mother and Child University Hospital Center.

% \textbf{Please add additional information for descriptive data such as:
% + What are the inclusion criteria for data selection?
% + Do we have any exclusion criteria?
% + 175 images from how many patients, how many girls, boys, and age range for all patients? }
% \textcolor{red}{Mario can you help me answer this question?}

These images were captured with a Microsoft Azure Kinect V2 camera. 
Every patient in the dataset had occlusions covering their body, even though most patients were photographed without their top garment. 
The occlusions ranged from bandages, cables, tubes, patches, and sheets to oxygen masks and respirators. 
The relative camera pose with regard to the patient varied from image to image.
In general, photos were most commonly taken from a forward-facing angle as far above the patient as possible, resulting in a $45^{\circ}$ angle between the bed, the patient, and the camera. 
The lighting inside of the ICU also varied from patient to patient, depending on their state of consciousness and the time at which the acquisitions were made. 
Naturally, we aim to collect the data with as little disturbance to the patient as possible. 
All lighting conditions in our dataset preserve the camera's working range of exposure and no photographs contain any saturation in high or low values at any point. 
Some patients could not physically be laid on the PICU bed. 
Because of this, the dataset contains images where the patient was sitting, sometimes on a wheelchair or a couch. 
For these acquisitions, the same camera angle was preserved. 
Finally, we split the dataset into three different parts where 80\% of the dataset was kept for training, 10\% of the dataset was used for validation and the remaining 10\% of the dataset was used for testing. 
Because of the ethical implications of our dataset, we cannot publicly release it; however, access to the database can be granted upon request by Dr. P. Jouvet.
%Please contact the PICU unit at CHUSJ for any data-related inquiries, including access to this dataset.

% \begin{figure}[htbp]
% \centerline{\includegraphics[width=\columnwidth]{Figures/ClassDistribution.png}}
% \caption{Distribution of occlusions instances in dataset}
% \label{fig:classdistribution}
% \end{figure}

\begin{figure}[!h]
\centerline{\includegraphics[width=0.9\columnwidth]{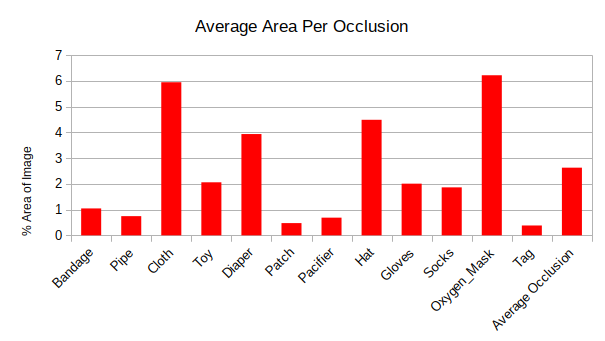}}
% \captionsetup{justification=centering}
\caption{Relative size distribution of various occlusion types. \\
 Each bar refers to each occlusion type averaged over all the images with that type of occlusion.
 The average occlusion displays the average area that any occlusion could occupy in an image. 
}

\label{fig:areadistribution}
\end{figure}

\subsection{Labeling}
We began the labeling of the images by defining a region of interest (ROI) located inside a bounding box where each patient's body is present. 
As previously mentioned, occlusions are defined as objects that may impede a direct view of a patient's body. 
For this reason, within the scope of our work, only occlusion segmentation within this defined ROI is considered.
The automatic detection of the patient's body is handled in a separate work. 
ROI-based analyses are quite common in the context of computer vision and are practically preferred to avoid feeding unnecessary noise and foreign objects such as machines, instruments, or other people into subsequent processing steps. 
After a careful review of the resulting dataset, we decided to label the most common obstructions that were found in our ROI. 
These were the following:
\begin{itemize}
    \item Bandages (both white and brown)
    \item Pipes (as well as cables and other long tubular plastic structures like nasal cannulas)
    \item Bedsheets (and pillows)
    \item Toys 
    \item Diapers
    \item Patches (ECG, EKG and others)
    \item Pacifiers
    \item Hats and tuques
    \item Gloves and mittens
    \item Socks and shoes
    \item Oxygen Masks
    \item Tags (for patient identification)
\end{itemize}

\begin{figure}[htbp]
\centerline{\includegraphics[width=0.8\columnwidth]{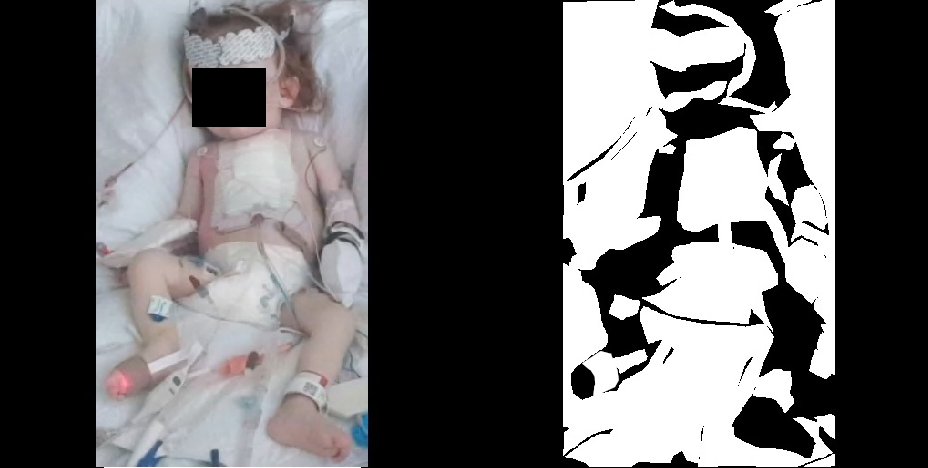}}
% \captionsetup{justification=centering}
\caption{A sampled image of a patient in the PICU with varied occlusions (left) and their corresponding
ground truth annotations (right) in our dataset.}
\label{fig:annotatedsample}
\end{figure}

Instance segmentation masks for occlusions were generated in the form of polygons. 
The different occlusion polygons were classified to have a better understanding of the most common occlusions in the PICU, as well as how much space each different occlusion occupies.
In Figure \ref{fig:areadistribution}, it is observed that the underlying distribution of different types of occlusions in the dataset is heavily unbalanced. 
Additionally, we can view the average area occupied by each type of occlusion in relation to the ROI of the image through the bar chart graph in Figure \ref{fig:areadistribution}. 
These severe imbalances complicate the fact that our training dataset is already small-sized, being detrimental for most common deep learning-based methods to effectively learn to segment semantically.
Taking this problem into account, \textit{within the scope of this work, we only consider a binary segmentation problem where each pixel in the image ROI is classified as either occlusion (class 1) or non-occlusion (class 0), as illustrated in Figure \ref{fig:annotatedsample}.}
Notably, despite our considered task being effectively binary segmentation, the multi-class instance segmentation labeling so far is still useful for the following augmentation stage. 

\subsection{Augmentation}
Although the considered segmentation task is binary, the effect of limited and unbalanced training data would still be problematic for the generalizability of most deep learning algorithms.
A common strategy to alleviate this negative impact is data augmentation, which is the modification of the original training data in a careful way such that more data can be utilized in the training process.
Data augmentation also allows us to mitigate data imbalance by overrepresenting minor classes, or in our case, by increasing the average area of occlusion per input image.

In this work, we use the following mild augmentations:
\begin{itemize}
    \item Flip ($50\%$ chance to be applied)
    \item Color Jittering ($20\% $ maximum change for brightness, contrast, saturation, hue and $50\%$ chance to be applied)
    \item Random Brightness and Contrast ($20\% $ maximum change for brightness and contrast)
\end{itemize}

\begin{figure}[htbp]
\centerline{\includegraphics[width=0.7\columnwidth]{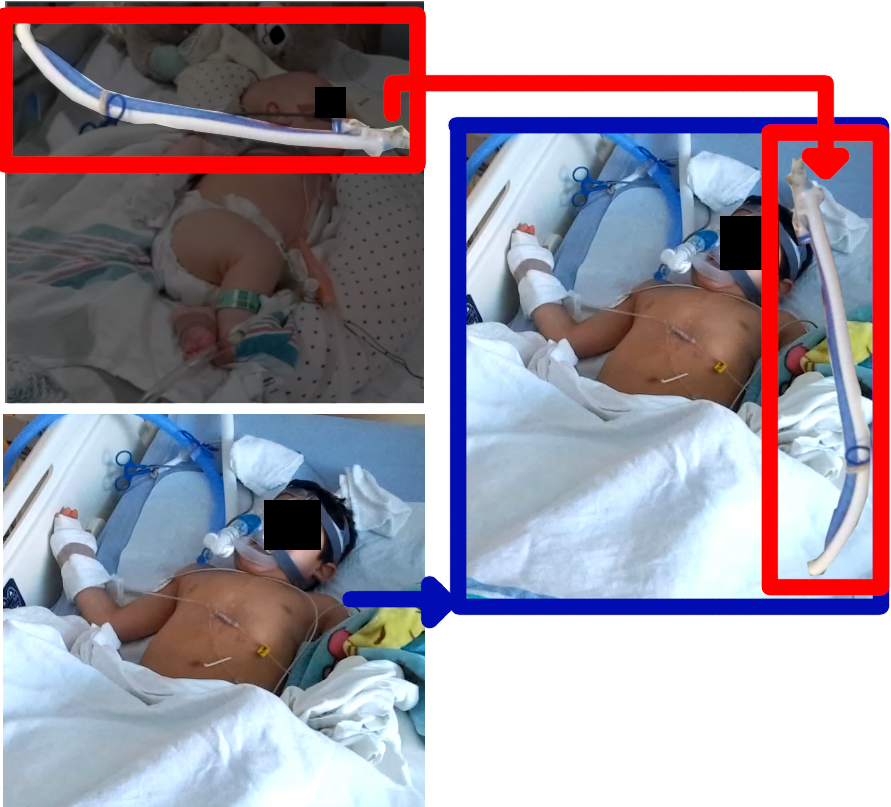}}
% \captionsetup{justification=centering}
\caption{An example of Copy-Paste augmentation on our dataset. (Top left) Source image.
(Bottom left) Target image. (Right) Occlusion transferred to the target image.}
\label{fig:copypaste}
\end{figure}

Nonetheless, these augmentations might not be sufficient since certain types of occlusions are very small or thin (such as pipes, patches...), as well as rarely appear in the dataset. 
Learning-based segmentation algorithms might struggle to learn to focus on these little details effectively. 
To handle this, we employ a kind of strong augmentation named CopyPaste \cite{ghiasi2021simple}. 
Its working principle is simple: we just need to copy both textural information and the associated segmentation mask of interested objects (occlusion ones in our case), geometrically transform them in simple ways and then transfer them to different positions or different images in the dataset  (Figure \ref{fig:copypaste}). 
Thanks to the annotated instance segmentation masks of individual occlusion objects, it is easy to implement CopyPaste.

On the other hand, using too many extreme transformations can also hurt the model's performance by introducing out-of-distribution as well as increasing unnecessary computational costs in pre-processing (\cite{he2019data}, \cite{wu2022knowledge}).
We were also careful not to include augmentations that would alter too much the final look of the test dataset, for example not  completely crowded with occlusions
We opted to Copy-Paste only $10\%$ of objects in each image.
We included two additional mandatory transformations for cropping and padding the image to prepare it for the model's 512 by 512 input size.

\subsection{Training Semantic Segmentation Network}
First, we need to train a neural network for our custom segmentation task.
We did not choose to train a vision transformer as they are difficult to train on very small datasets \cite{zhu2023understanding} and computationally more demanding than regular CNN-based networks \cite{steiner2021train}\cite{deininger2022comparative}.  
Additionally, because of the sensitive nature of the dataset at hand, we could not simply export the data from the hospital's servers and compute the training in the cloud, therefore we opted to reduce the computational complexity of the training. 
Particularly, the DeepLabV3+ model is chosen since it is a well-recognized CNN-based semantic segmentation model.
For the purpose of our experiments, we denote this model as our semantic segmentation model ($\mathcal{M}_\text{SEM}$). 
To speed up the training process, we initialized the network with pre-trained COCO \cite{lin2014microsoft} dataset weights.
To adapt the architecture to our segmentation task, we retained the feature extraction layers while attaching a new classifier head that outputs 2 classes instead, following conventional fine-tuning works \cite{tajbakhsh2016convolutional}.
Although weight freezing \cite{brock2017freezeout} can be recommended for fine-tuning or transfer learning, picking which layers to freeze is often formulated as a heuristic, resource-intensive process \cite{herrera2022optimal}. 
Additionally, we judged the features at higher levels in the network that were learned from the COCO dataset to be far different from the ones in our dataset.
Thus, we opted not to freeze any weights at any layer in our model in this work.

For the training of our $\mathcal{M}_\text{SEM}$ network, we opted to feed our dataset in batches of 8.
Every time that the model was fed an image from the dataset, transformations were applied to each image and label as the training took place.
We trained for $122$ epochs (terminated by early stopping), tracking the performance metrics of the model on the validation dataset after each epoch.
We opted to use a learning rate scheduler based on cosine annealing with warm restarts\cite{loshchilov2016sgdr}.
This was done in order to augment the probability of leaving a local minimum during our optimization and exploring a broader search space.
After the hyperparameter-tuning procedure on the validation set, we finally employed a learning rate of $10^{-2}$ with a weight decay of $10^{-4}$ and a learning momentum of $0.9$. 
For the cosine annealing with warm restarts, $T_{0}$ of $50$ was chosen as well as a $T_{mult}$ of $1$ which means there is only one warm restart every period. 
In order to limit the learning rate above a threshold so as to not waste computing power, an $ETA_{min}$ learning rate of $5 \times 10^{-4}$ was also applied.

\begin{figure}[htbp]
\centerline{\includegraphics[width=0.95\columnwidth]{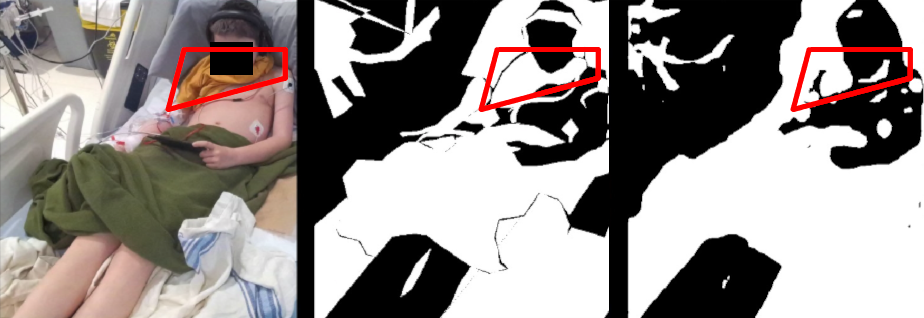}}
% \captionsetup{justification=centering}
\caption{An example of $\mathcal{M}_\text{SEM}$'s prediction. (Left) Input image. (Middle) Occlusion segmentation annotation. (Right) Occlusion segmentation mask by $\mathcal{M}_\text{SEM}$. The scarf (highlighted in the red box) is roughly localized and identified as occlusion. However, the predicted mask shape of the scarf is inaccurate.}
\label{fig:semexample}
\end{figure}

\subsection{Prediction Enhancement with SAM} \label{subsection:enhancement}
Although the trained $\mathcal{M}_\text{SEM}$ visually demonstrates its potential to recognize and localize occlusions, the shape of recognized occlusions can be inaccurate (Figure \ref{fig:semexample}).
This is expected since:
\begin{itemize}
    \item The training dataset is limited in size, and
    \item Certain types of occlusions are small and/or have thin shapes, while the cropped input image is noisy and of low resolution.
\end{itemize}
To overcome these limits, we need a more powerful transfer learning algorithm that can generalize to this uncommon domain. 
Towards this end, we leveraged the Segment-Anything-Model (SAM), a state-of-the-art transformer-based image segmentation model with strong zero-shot learning capability. 

SAM can be promptable which means we can let it focus on specific regions of an image by using bounding boxes, points, or text as prompts.
However, we decided not to use bounding boxes as many occlusions in our PICU setting such as cables and tubes are long, overlap with other occlusions, and have no clearly defined shapes.
On the other hand, our initial experimentation with SAM revealed that its usage with single points and text prompts did not necessarily achieve robust segmentation.
Therefore, an automatic method for prompting was used, similar to the one presented in \cite{kirillov2023segment} with a grid of $40 \times 40$ points for prompting SAM.
This increased point density is due to the resolution of the images at hand and the small nature of some occlusions. 
In order to limit incomplete and less-than-satisfactory regions, we set a confidence threshold of $90\%$. 
The stability score of the model was set to $85\%$. 
In order to avoid segmenting very small regions in the image that are not related to any occlusions, we also used a minimum area threshold of $5$ pixels for any SAM segmentation mask.

In our experiments, SAM can offer high-quality image segmentation and produce "superpixels" which contain pixels of uniform texture and contextual information and at the same time have accurate boundaries. 
However, it has certain known limits:
\begin{itemize}
    \item Each image segment is not explicitly associated with a specific semantic class.
    \item Each image segment might only represent a part of an object. 
\end{itemize}
To tackle this problem, we opted to exploit the strengths of both the SAM model (denoted as $\mathcal{M}_\text{SAM}$) and $\mathcal{M}_\text{SEM}$ predictions by fusing them through our proposed soft voting mechanism. 
We named this fusion mechanism SAM-powered Occlusion Segmentation via Soft-voting (SOSS).
A simplified diagram of the proposed pipeline is presented in Figure \ref{fig:pipeline}.

Specifically, our pipeline consists of two deep neural networks.
The first (DeepLabV3+-based) network is responsible for segmenting the input image and generating a coarse semantic mask proposal for occlusion. 
The second (SAM-based) one provides fine-grained partitioning without associating it with semantic labels. 
These two types of masks are then combined using our confidence-based soft-voting approach to produce a final, improved occlusion segmentation mask.\\

\begin{assum} \label{assum:assumption1}
    For any pair of pixel $i^\text{th}$ and $j^\text{th}$ of the same input image (I), with their respective SAM output denoted as $\mathcal{M}_\text{SAM}^{(i)}$ and $\mathcal{M}_\text{SAM}^{(j)}$, and their hidden occlusion class respectively denoted as $o^{(i)}$ and $o^{(j)}$, it is assumed that:
    \begin{equation}
        \mathcal{P}\left(o^{(i)} = o^{(j)} | \mathcal{M}_\text{SAM}^{(i)} = \mathcal{M}_\text{SAM}^{(j)} \right) = 1
    \end{equation}
\end{assum}
\vskip 3mm

Under this assumption, any pair of points belonging to the same segment by SAM should have the same occlusion class. 

This implies that SAM segments can be regarded as the finest granularity for semantic segmentation. 
In fact, this assumption is supported by the observations made by the original paper \cite{kirillov2023segment} for object proposal tasks, instance segmentation tasks, as well as SAM's potential to carry semantic information in the latent embedding space without explicit semantic supervision. 
This assumption is very powerful since it allows us to effectively constrain the semantic relationship of neighboring pixels and integrate SAM into more complicated tasks in a flexible fashion.

Based on this assumption, we then proposed our fusion algorithm (SAM-powered Occlusion Segmentation via Soft-voting, or SOSS for short) as follows:
\begin{algorithm}[H]
\caption{SAM-powered Occlusion Segmentation via Soft-voting (SOSS)}\label{alg:alg1}
\begin{algorithmic}
\STATE 
\STATE {\textsc{PREDICT}}$(\mathbf{\mathcal{M}_\text{SEM}}$\hspace{0.2cm}$ \mathbf{\mathcal{M}_\text{SAM}}$\hspace{0.2cm}$\mathbf{I})$
\STATE \hspace{0.5cm}$ \mathbf{j} = (1,1)$
\STATE \hspace{0.5cm}$ \mathbf{P_{SAM}} \gets \mathbf{\mathcal{M}_\text{SAM}}(I) $
\STATE \hspace{0.5cm}$ \mathbf{P_{SEM}} \gets \mathbf{\mathcal{M}_\text{SEM}}(I) $
\STATE \hspace{0.5cm}$ \forall $\hspace{0.2cm}$ \mathbf{S} \in \mathbf{P_{SAM}}:$
\STATE \hspace{0.75cm}$ P_{mod} \gets P_{SEM}S$
\STATE \hspace{0.75cm}$ C \gets \arg \max (P_{mod}j)$
\STATE \hspace{0.75cm}$ \mathbf{P_{FINAL}} \gets C\mathbf{S}+\mathbf{P_{FINAL}}$
\STATE \hspace{0.5cm}\textbf{return} $\mathbf{P_{FINAL}}$
\STATE 
\end{algorithmic}
\label{alg1}
\end{algorithm}

Using these settings, any image ($I$) of size $H\times W$ can be used by our $\mathcal{M}_\text{SAM}$ to infer a binary prediction matrix ($P_{SAM}$). Given a total number of $O$ separate objects detected by $\mathcal{M}_\text{SAM}$, $P_{SAM}$ will be of $(H,W,O)$ dimensions. Therefore, inside $P_{SAM}$, there are $O$ binary 2-dimensional matrices ($B$), each representing a "patch" (equivalently, "superpixel" or segment proposal) which corresponds to the filled silhouette of different objects in the image. 
These $O$ matrices are believed to have better edges than the semantic predictions of $P_{SEM}$ after applying any given threshold.
In order to leverage the classification aspect of our $\mathcal{M}_\text{SEM}$ and the sharp segmentation masks of our $\mathcal{M}_\text{SAM}$, we used a soft-voting algorithm.
This algorithm attributes a class $C$ from $N$ different classes to each matrix $B$ in $P_{SAM}$. 
In order to calculate which class to attribute, we mask $P_{SEM}$ along its $H$ and $W$ dimensions using every matrix $B$ in $P_{SAM}$.
The resulting masked prediction ($P_{mod}$) is used to determine the class C by locating the index of the maximum semantic confidence value.
To do so, we sum $P_{mod}$ along its $N$ dimension, resulting in a uni-dimensional vector of length $N$.
The index of the maximum value in this vector corresponds to the final occlusion class $C$. 

The intuition behind this algorithm is that if the segmentation by SAM is accurate, we can query all the pixels belonging to each segment for their occlusion classes which are given by $\mathcal{M}_\text{SEM}$.
The class with the highest votes becomes the final prediction for individual segments.
Taking into account the uncertainty of $\mathcal{M}_\text{SEM}$'s predictions, especially when being trained on a small dataset, we propose to use soft-voting, instead of hard voting, to address the overconfidence problem with $\mathcal{M}_\text{SEM}$ and take more conservative polling.
This fusion mechanism can help overcome the aforementioned problems of SAM in that if the majority of the pixels in a pair of segments (by SAM) believe that they belong to the same class, the two segments can be joined into a single segment with an associated semantic class. 
It can also improve $\mathcal{M}_\text{SEM}$'s prediction since each final semantic segment is formed by a union of SAM's segments and thus is supposed to possess more accurate boundaries.

\section{Experiment Results and Discussion}
\label{sec:results}

% \begin{figure}[!h]
% \centerline{\includegraphics[width=\columnwidth]{Figures/Results.png}}
% \caption{Results on our test set from the DeepLabV3+ model and the combined DeepLabV3+ model with SAM fusion }
% \label{fig:results}
% \end{figure}

\begin{figure*}[!t]
\centerline{\includegraphics[width=0.8\textwidth]{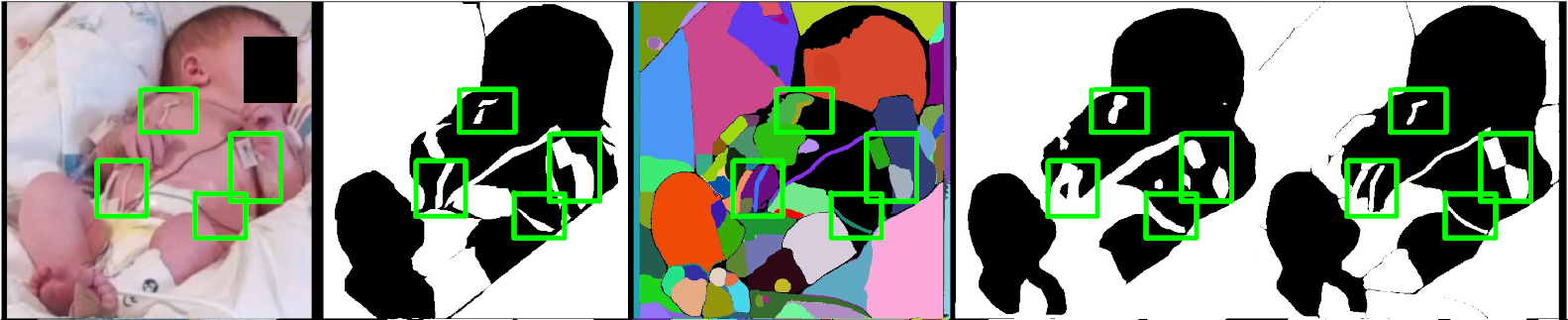}}
\centerline{}
\centerline{\includegraphics[width=0.8\textwidth]{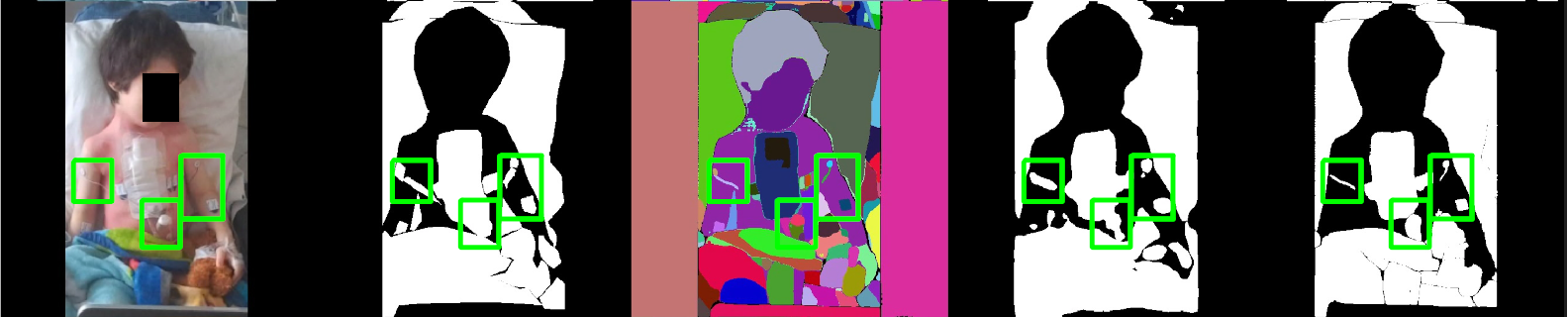}}
\centerline{}
\centerline{\includegraphics[width=0.8\textwidth]{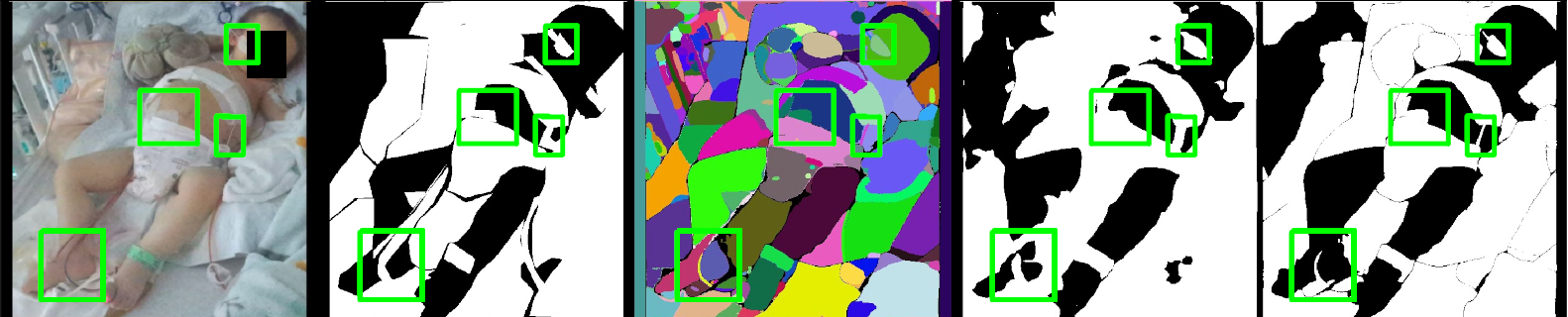}}
\centerline{}
\centerline{\includegraphics[width=0.8\textwidth]{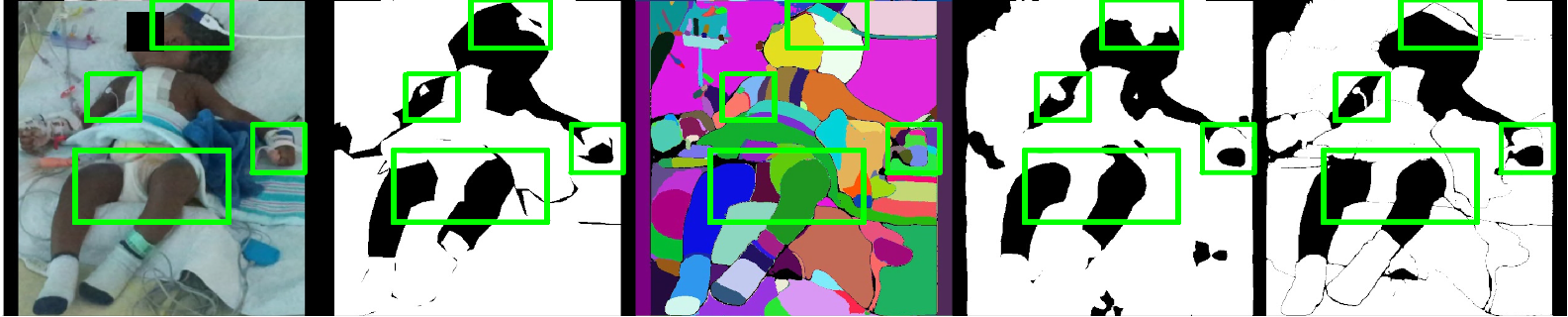}}
% \captionsetup{justification=centering}
\caption{Improved predictions on the held-out test data. From left to right: input images ($1^\text{st}$ column), 
ground truths ($2^\text{nd}$ column), SAM masks ($3^\text{rd}$ column), predictions from $\mathcal{M}_\text{SEM}$ ($4^\text{th}$ column), and final binary occlusion segmentation mask ($5^\text{th}$ column). 
In the $2^\text{nd}, 4^\text{th}$ and $5^\text{th}$ columns, white regions are classified as occlusions while black ones are classified as non-occlusions. In the $3^\text{rd}$ column, different colors simply indicate different image segments without associated occlusion class, which is the main shortcoming of SAM in our task. Green boxes highlight occlusion easily visible segmentation improvements.} \label{fig:prediction_success}
\end{figure*}
% \begin{figure*}[!t]
% \centerline{\includegraphics[width=0.8\textwidth]{Figures/ex5.png}}
% \caption{Result on a test data point from the proposed pipeline. This image shows how specific patch-shaped occlusions can be refined through our pipeline, as well as some ambiguous background elements.}
% \label{fig:predictions2}
% \end{figure*}

\begin{figure*}[!t]
\centerline{\includegraphics[width=0.8\textwidth]{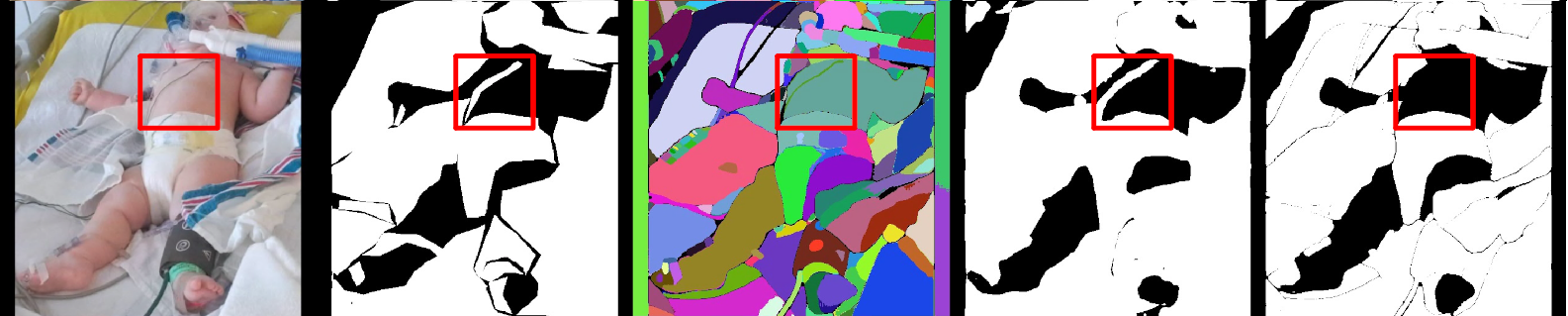}}
\centerline{}
\centerline{\includegraphics[width=0.8\textwidth]{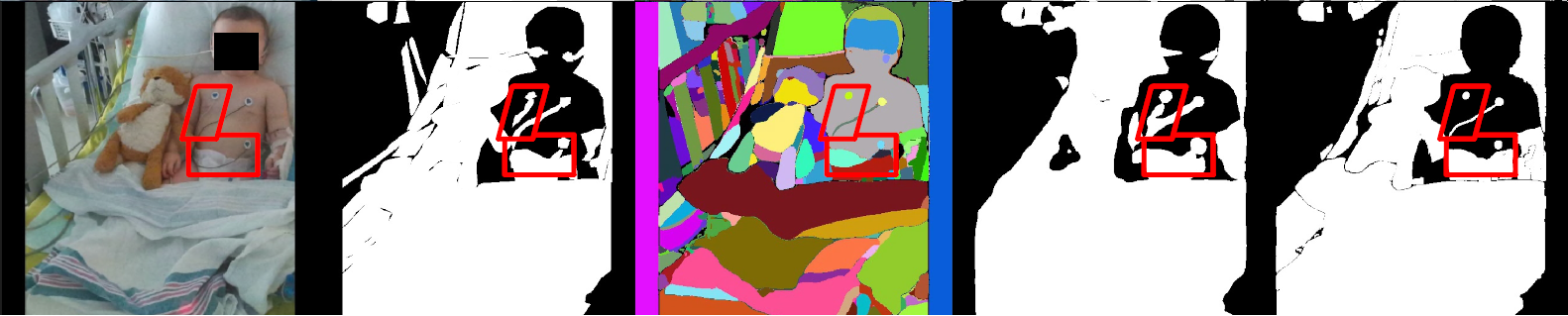}}
% \captionsetup{justification=centering}
\caption{Failure examples of SOSS. From left to right: input images ($1^\text{st}$ column), ground truths ($2^\text{nd}$ column),
SAM masks ($3^\text{rd}$ column), predictions from $\mathcal{M}_\text{SEM}$ ($4^\text{th}$ column), and final binary
occlusion segmentation mask ($5^\text{th}$ column). In the $2^\text{nd}, 4^\text{th}$ and $5^\text{th}$ columns, white regions are classified as occlusions while black ones are classified as non-occlusions. 
In the $3^\text{rd}$ column, different colors simply indicate different image segments without associated occlusion class, which is the main shortcoming of SAM in our task. Red boxes highlight occlusion segmentation failures.}\label{fig:prediction_failure}
\end{figure*}
A variety of tools have been used to implement our proposed pipeline.
To annotate the dataset we used the LabelStudio \cite{LabelStudio} toolset. 
The training of our $\mathcal{M}_\text{SEM}$ network was done using Pytorch\cite{paszke2017automatic}. The data was also augmented using the Albumentations \cite{2018arXiv180906839B} library, besides the CopyPaste augmentation that was made to fit as another transform in the same pipeline. 
We also leveraged OpenCV\cite{itseez2015opencv} and Numpy \cite{harris2020array} for image manipulation.
To evaluate the pipeline, a held-out test set consisting of $10\%$ of the total images was used without any augmentations. 

\subsection{Quantitative results}
To alleviate the effect of class imbalance which can be strong in our dataset, we included the following metrics for performance comparison:
\begin{itemize}
    \item Accuracy: $Acc=\frac{TP+TN}{TP+TN+FP+FN}$
    \item Precision: $Prec=\frac{TP}{TP+FP}$
    \item Recall: $Rec=\frac{TP}{TP+FN}$
    \item F1 score: $F_1=(\frac{0.5}{Prec}+\frac{0.5}{Rec})^{-1}$
    \item Intersection over Union (IoU): $IoU= \frac{|True \cap Predicted|}{|True \cup Predicted|}$
    \item Area Under the receiver operating characteristic Curve (AUC)
\end{itemize}
where F and T stand for False and True; P and N stand for Positive and Negative, respectively. Furthermore, TN and TP stand for true negative and true positive respectively, and are the number of negative and positive cases correctly classified. FP and FN represent false positives and false negatives and the number of incorrectly predicted positive and negative cases.
These metrics are per class; in the following discussion, we report them for the occlusion class (class 1).

Specifically, Table \ref{tab:results} shows the occlusion segmentation performance of our $\mathcal{M}_\text{SEM}$, and then those attained by the whole proposed SOSS pipeline.
We can see that the proposed method consistently improves the predictions across all metrics, with an average of $2.75\%$ gain in performance.
It should be noted that if we focused on the occlusions in the foreground only and excluded the ones on the patient's side, the performance gain could have been bigger.
Overall, this increase clearly demonstrates the effectiveness of our proposed pipeline in the intended task of occlusion segmentation.

\begin{table}[!h]
\centering
\caption{Performance of CHU-SJ artifact segmentation task}
\label{tab:results}
\begin{tabular}{c|cc}
\hline
Metrics           & DLV3+& SOSS (ours)   \\ \hline
$Accuracy \uparrow$         & 89.1 & \textbf{92.5}  \\
$F1 \uparrow$               & 90.0 & \textbf{92.0}  \\
$Precision\uparrow$         & 88.1 & \textbf{90.3}  \\
$Recall \uparrow$           & 92.0 & \textbf{93.8}  \\
$IoU \uparrow$              & 81.8 & \textbf{85.2}  \\
$AUC \uparrow$              & 88.9 & \textbf{92.6}  \\ \hline
\end{tabular}
\end{table}

\subsection{Qualitative results}
To better understand the improvements attained by the proposed pipeline, Figure \ref{fig:prediction_success} and \ref{fig:prediction_failure} present images sampled from our test set and the corresponding predictions at different steps of our pipeline. 

By visually comparing the three right-most columns with the input images ($1^\text{st}$ column) and the occlusion mask annotation ($2^\text{nd}$ column), we can evaluate how good the segmentation masks are at each step. \\
For instance, the second row in Figure \ref{fig:prediction_success} demonstrates a pediatric patient in PICU. 
In each image of this second row, the left-most green box focuses on a fine cable attached to his right arm, with the corresponding occlusion annotation in the $2^\text{nd}$ column. 
The $3^\text{rd}$ column provides an accurate, fine-detailed instance segmentation of this cable by SAM; however, this segment is only associated with an integer without any occlusion classification, which might be confusing for our task.
On the contrary, the $4^\text{th}$ column demonstrates $\mathcal{M}_\text{SEM}$'s capability of assigning occlusion class to this segment. 
Although the occlusion is correctly detected here, the shape of this segment is rather coarse and bigger than the original cable in the input image.
Finally, the $4^\text{th}$ (SOSS's prediction) correctly captures both the segment's semantic label (occlusion-in white) as well as the accurate shape of this cable. \\
Overall, close visual inspection in Figure \ref{fig:prediction_success} reveals that $\mathcal{M}_\text{SEM}$ ($4^\text{th}$ column) is relatively accurate about occlusion localization despite being trained on a limited amount of data.
Surprisingly, it shows consistent performance with complex illumination conditions, diverse patients' ages and skin colors, cluttered backgrounds, and particularly with a high level of occlusion density. 
The segmentation boundaries seem somewhat stable (less wiggly and wavy) with signs of slight over-smoothing.
The most unpleasant things about $\mathcal{M}_\text{SEM}$ are the frequent overrepresentation of small and thin objects, as well as occasional straying spots (Figure \ref{fig:prediction_success}, top row).\\
On the other hand, SAM's predictions ($4^\text{th}$ column) produce highly accurate image segmentation in which the segments' edges and those in the input image agree to an excellent extent.
As mentioned in Subsection \ref{subsection:enhancement}, SAM's main drawback lies in its lack of semantic label assignment.
For example, even though we can observe an accurate segment of what is seemingly a cable, we do not know if that segment can be classified as an occlusion or not.

Finally, by comparing the segmentation masks of SOSS ($5^\text{th}$ column) with the other columns, it is observed that the segmentation quality is greatly improved in the following aspects:
\begin{itemize}
    \item Fine details of thin objects (pipes) or small ones (patches) strongly agree with the ground truth. 
    Large objects predicted as whole regions can be subdivided into smaller instances that possess different textures or colors.
    It is worth noticing that in many cases, the final segments even look more faithful to the raw input image than the annotated masks.
    If the annotation were done at higher resolutions, the reported performance gain of SAM refinement could even be bigger.  
    \item Not only have coarse details been refined, but also missing details in $\mathcal{M}_\text{SEM}$ can be recovered (such as in the first image of Figure \ref{fig:prediction_success}).
    Regions that are spotty in the classifier's prediction can be rendered whole again after fusion. 
    Additionally, small objects such as electrodes, patches, and tags are preserved.
\end{itemize}

However, we do realize occasional failure cases of SOSS from Figure \ref{fig:prediction_failure}, typically with thin, small objects as cables.
For example, the red box in the top row of Figure \ref{fig:prediction_failure} shows a correct recognition of a cable by $\mathcal{M}_\text{SEM}$ ($2^\text{nd}$ column). 
However, in the final predictions by SOSS ($5^\text{th}$ column), the corresponding mask disappears.
The same can be said about the cables lying on the right side of the chest and the left side of the abdomen of the patient in the second row.
It can be seen that in those cases, SAM is not aware of the cables, thus Assumption \ref{assum:assumption1} does not hold.
This is a shortcoming when the pipeline is overconfident about SAM's predictions.

\section{Conclusion}
In this paper, we proposed a pipeline for efficiently segmenting occlusions in a clinical setting with little data by leveraging pre-trained semantic segmentation models, data augmentation, and mature promptable segmentation models like SAM.
Our findings suggest that efficient segmentation of occlusions in a PICU setting is a task that can be accomplished with limited data and the help of strong zero-shot segmentation models.

On the other hand, there are certain shortcomings associated with our methodology, mostly associated with SAM:

\begin{itemize}
    \item Slow inference: The main bottleneck in our pipeline is SAM, which takes significantly longer time for inference compared to $\mathcal{M}_\text{SEM}$. 
    
    \item Prompting resolution: Masks generated by SAM can be quite sensitive to the chosen prompting resolution.
    SAM can skip objects because of their small or thin size, as in the case of cables that perhaps fall between prompting points in the image.
   Generally, more prompted points require higher computing resources.
    This is however limited by the computational power of the existing hospital's dedicated server; in other words, sufficiently fine-grained point prompting might not be attainable.
    \item Overconfidence in SAM: Our proposed pipeline is based on the assumption of SAM accuracy.
    Even though SAM's predictions are very accurate in general, there are certain cases where SAM is unaware of very small, irregular objects.
\end{itemize}

Therefore, our future research can be extended in the following directions:
\begin{itemize}
    \item Utilizing SAM in a more efficient way:
    For example, we can optimize the prompting resolution to balance between accuracy and inference speed.
    Another possible improvement is to use faster implementation of SAM, for example with a recently proposed architecture in \cite{zhao2023fast}.
    \item Leveraging multi-modality:
    Additional synchronized modalities such as depth images and thermal images captured by the hospital acquisition system can be provided for our dataset. 
    SAM is also reported to work with depth modality \cite{cen2023sad} and thus can exploit rich geometric information from depth images aside from textural information given by RGB images, which might further boost the segmentation performance.
\end{itemize}

\section{Acknowledgement}
Clinical data were provided by the Research Center of CHU Sainte-Justine Hospital, University of Montreal. This work was supported in part by the Natural Sciences and Engineering Research Council (NSERC), in part by the Institut de Valorisation des données de l’Université de Montréal (IVADO), and in part by the Fonds de la recherche en sante du Quebec (FRQS).

\bibliographystyle{IEEEtran}
\flushend
\bibliography{IEEEabrv,Bibliography}
  
\end{document}